# WeatherNet: Recognising weather and visual conditions from street-level images using deep residual learning

Mohamed R. Ibrahim[1], James Haworth[2] and Tao Cheng[3]
Department of Civil, Environmental and Geomatic Engineering, University College London (UCL)
[1]mohamed.ibrahim.17@ucl.ac.uk, [2]j.haworth@ucl.ac.uk, [3]tao.cheng@ucl.ac.uk

*Abstract*—Extracting information related to weather and visual conditions at a given time and space is indispensable for scene awareness, which strongly impacts our behaviours, from simply walking in a city to riding a bike, driving a car, or autonomous drive-assistance. Despite the significance of this subject, it is still not been fully addressed by the machine intelligence relying on deep learning and computer vision to detect the multi-labels of weather and visual conditions with a unified method that can be easily used for practice. What has been achieved to-date is rather sectorial models that address limited number of labels that do not cover the wide spectrum of weather and visual conditions. Nonetheless, weather and visual conditions are often addressed individually. In this paper, we introduce a novel framework to automatically extract this information from street-level images relying on deep learning and computer vision using a unified method without any pre-defined constraints in the processed images. A pipeline of four deep Convolutional Neural Network (CNN) models, so-called the WeatherNet, is trained, relying on residual learning using ResNet50 architecture, to extract various weather and visual conditions such as Dawn/dusk, day and night for time detection, and glare for lighting conditions, and clear, rainy, snowy, and foggy for weather conditions. The WeatherNet shows strong performance in extracting this information from user-defined images or video streams that can be used not limited to: autonomous vehicles and drive-assistance systems**,** tracking behaviours, safety-related research, or even for better understanding cities through images for policy-makers.

1. INTRODUCTION

Cities are complex entities by nature due to the multiple, interconnected components of their systems. Features of the physical environment extracted from images, or so-called urban scenes, have great potential for analysing and modelling cities because they can contain information on a range of factors, such as people and transport modes, geometric structure, land use, urban components, illumination and weather conditions (Narasimhan et al., 2002). In recent years, computer vision techniques have shown progress in extracting and quantifying these features (Ibrahim et al., In press, 2019).

This article is concerned with the recognition of weather and visual conditions, which are two related but separate aspects of urban scenes that can be extracted in order to better understand the dynamics of the appearance of the physical environment (Liu et al., 2016). In this study, we refer to visual conditions as the significant changes of the appearance of cities during dawn/dusk, day or night-time, including the effect of glare on visibility, whereas weather conditions as the meteorological changes of the environment due to precipitation including clear, rainy, foggy or snowy weather. They represent crucial factors for many urban and transport-related studies, in which they show significant associations when it comes to traffic surveillance and tackling behavioural and safety-related research (Lacherez et al., 2013). For example, walking, cycling or driving in rainy weather is associated with a higher risk of experiencing an incident than in clear weather (Branion-Calles et al., 2017; Lacherez et al., 2013). Not only rain conditions, but also fog, snow, or even glare can cause a risk when it comes to moving from one place to another (Branion-Calles et al., 2017; Dozza and Werneke, 2014). Importantly, it is not only the inherent risk that different weather and visual conditions pose to human life that is of interest to researchers. Scene awareness for autonomous navigation in cities is highly influenced by the dynamics of weather and visual conditions and it is imperative for any vision system to cope with them simultaneously (Narasimhan, 2003). For example, object detection algorithms must perform well in fog and glare, as well as clear conditions, in order to be reliable. Accordingly, finding an automatic approach to extract this information from images or video streams is in high demand for computer scientists, planners and policy-makers.

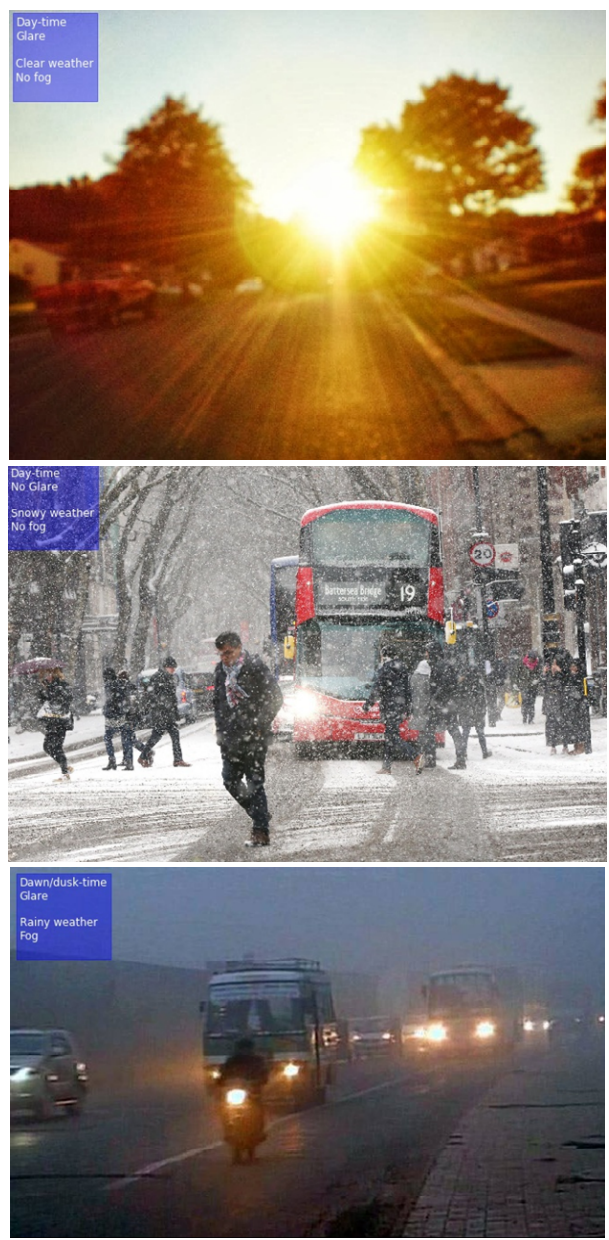

Figure 1: WeatherNet Output

While there are different methods that are used to understand the dynamics of weather and visual conditions, a knowledge gap appears when addressing this subject. To date, these two crucial domains - weather and visual conditions - have been studied individually, ignoring the importance of understanding the dynamics and impact of one domain on the other. There is no unified method that can extract information related to both weather and visual conditions from a street-level image that can be utilised by planners and policy-makers.

Building on the current advances of scene awareness based on computer vision, in this paper, we present a novel framework, WeatherNet, that aims to recognise and map the dynamics of weather and visual conditions with a unified method. The framework takes single-images as input, and does not require pre-defined constraints such as the camera angle, area of interest, etc. The WeatherNet relies on multiple deep Convolutional Neural Network (CNN) models that aim to recognise visibility related conditions such as dusk/dawn, day or night-time, glare, and weather conditions such as clear, fog, cloud, rain, and snow. The motivation of the WeatherNet is to practically extract and map weather information in cities that could help planners and policy-makers to analyse cities, and to contribute to the intelligent systems of navigation in cities and autonomous driving. Figure 1 shows the output of WeatherNet framework.

After the introduction section, the paper is structured in four sections: first, in section 2 we review related work, discussing the methods used and their limitations. Second, in section 3 we introduce the materials and methods used for the WeatherNet. Third, in section 4, we show the results of the different CNN models and their diagnostics, discussing the outcome and the current limitations. Some discussions are provided in section 5, before we draw conclusions and present our recommendations for future work in section 6.

## 2. RELATED WORK

Various weather and visual conditions have been detected relying on a wide spectrum of computer vision algorithms. Here, we categorised them into four broad types: Mathematical models, filter-based models, machine learning models using shallow algorithms and deep models using a convolution structure.

### 2.1 Mathematical Models

Mathematical models for weather detection have focussed mainly on fog detection, with applications in drive-assistant navigation systems. For instance, (Lagorio et al., 2008) developed a statistical framework based on the mixture of Gaussians to detect binary weather conditions, snow and fog, based on the dynamics of the spatial and temporal dimensions of images. The method for rain detection senses the moving textures of rain due to the transparency of water drops relative to light. This approach requires settings for capturing images to be known, such as camera optics and the viewing distance, etc. Therefore, although the results for detecting snow and fog are promising, the method can only be applied in specific, controlled cases and is insufficient for capturing different weather events. A method based on Koschmieder's law, (Middleton, 1952) was developed to detect fog in daytime and estimate visibility distance from images, in which sky and road are present, based on the theory of how the apparent luminance of an object is observed against the background (i.e. sky) on the horizon (Hautiére et al., 2006). Apart from the achieved accuracy, this method is limited for detecting dog during day-time only. Another model is developed to detect fog based on Canny Edge detection algorithm (Negru and Nedevschi, 2013). While the model is capable of estimating not only fog but also the visibility distance from black-and-white images, the accuracy of the model reduces when analysing urban scene images crowded with vehicles. Although the method shows good potential in detecting fog from daytime images, given the nature of the model algorithms, the proposed method is limited to fog detection in constraint conditions such as daytime and requires further development to detect fog at night.

### 2.2 Filtering-based models

Moving towards filtering images, different techniques have been achieved to recognise weather conditions based on their visual characteristics and features. For example, (Li Shen and Ping Tan, 2009) developed a model to detect weather conditions – such as sunny and cloudy weather - based on the global illumination, relying on the association between scene illumination and weather conditions. (Yan et al., 2009) present an algorithm for weather recognition from images relying on information such as; a road, a histogram of colour and gradient amplitude. (Pavlic et al., 2012) developed a model to detect dense fog from black-and-white daytime images relying on Gabor filters represented in different scales, orientations and frequencies. (Cord and Aubert, 2011) developed a model to detect raindrops based on the photometric characteristics of raindrops relying on variations of the gradient in the image.

The purpose or the accuracy for these individual models may vary. However, the common limitations of these models that they all require pre-defined settings for the models to function. Such settings are often limited for a given purpose or task, i.e. fog detection. This reduces the ability of the given models to be transferable to tackle the other aspect of weather and visual conditions.

### 2.3 Machine Learning Models

Machine learning models have shown progress in recognising the multi-class conditions of weather. For example, (Roser and

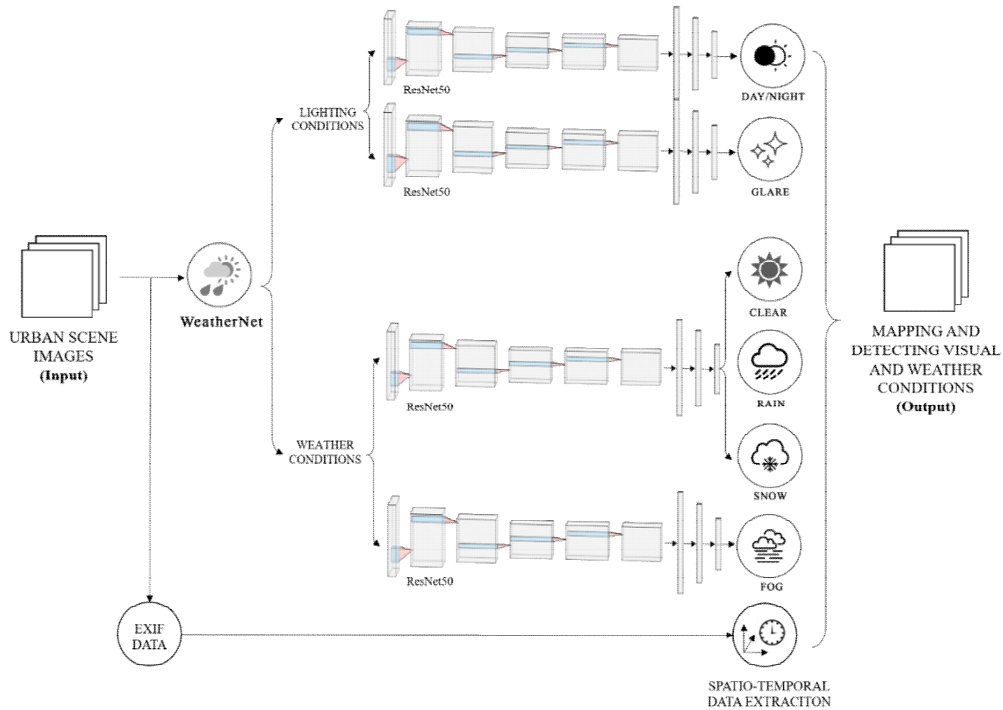
Figure 2: The framework of the WeatherNet

Moosmann, 2008) developed a model to classify weather conditions using a Support Vector Machine (SVM) trained on single colour images. However, the model is limited to detecting three weather conditions; clear weather, light, and heavy rain. (Chu et al., 2017) used a Random Forest classifier to classify weather conditions using tagged images of weather data such as weather condition (sunny, foggy, cloudy, etc), temperature, and humidity. Lastly, (Zhang et al., 2016) developed a method to classify weather conditions based on a general framework that aims to extract multiple features such as sky, rain, snowflakes, shadow, dark channel, saturation and contrast relying on k-nearest neighbours and SVM. Still, the achieved models are rather confined for a given sector of weather classification, in which their methods cannot be transferred to cover the different classes of weather and visual conditions.

### 2.4 Deep Learning Models

Computer vision relying on deep models, specifically, CNN models, has shown progress for image processing tasks and scene awareness (LeCun et al., 2015). A range of applications based on classifying, segmenting and localising pixels from street-level images has become a common approach for understanding the various components of an urban scene (Chaurasia and Culurciello, 2017; Li et al., 2017; Yang et al., 2018; Zhou et al., 2017). Similarly, various models have been developed to classify weather from features extracted based on a convolution structure of deep models. For instance, a CNN model coupled with sparse decomposition is trained to classify weather conditions (Liu et al., 2017). Also, a binary CNN model is trained to classify images as either cloudy or sunny (Elhoseiny et al., 2015; Lu et al., 2017). However, this model remains limited to the given binary classes of weather, ignoring the complexity of the addressed subject. Building on the previous methods, (Guerra et al., 2018) developed a framework relying on super-pixel masks, CNN and SVM classifiers to detect three weather classes; rain, fog, and snow. While this model shows progress in recognising more weather classes, it only sees weather conditions as exclusive classes, ignoring the co-existence of two or more classes in a single image for a given time. Lastly, to solve the combination issue of the existence of multiple weather class in a single image, Zhao et al. (2019) used a CNN based model that includes an attention-layer to allow the model to infer more than a class for a given time depending on the characteristics of the input image. While this model shows progress in classifying multiple weather conditions and their combinations (sunny, cloudy, foggy, rainy, and snowy), it still ignores the dynamics of visual conditions and the time of day that may influence weather classification accuracy.

### 2.5 Summary

Based on current literature, there is still on-going research to cover the current limitation in addressing the weather and visual conditions simultaneously, in which addressing one domain only would not necessarily cover the dynamics of the appearance of urban scenes. For instance, cities may appear darker when it rains in the day-time than during clear weather at the same time. While the above-mentioned models show progress in the given tasks, there are number of knowledge gaps that needs to be addressed to cover the stated subject of weather and visual classification, which are: 1) these crucial domains- weather and visual conditions- have been studied individually, ignoring the importance of understanding the dynamics and impact of one domain on the other. There is no unified method that can extract information related to both weather and visual conditions from a street-level image; 2) weather classification has been treated with a limited number of labels, ignoring the variation of weather conditions. Even when weather is treated as a multi-label classification, a knowledge gap appears in representing scenes with multiple labels that simultaneously co-exist; 3) current models used to classify weather and visual conditions are either limited to a presenting requirement or limited in accuracy. The methods are not up-to-date with the state-of-the-art of machine vision research (i.e. no models rely on residual learning to understand weather).

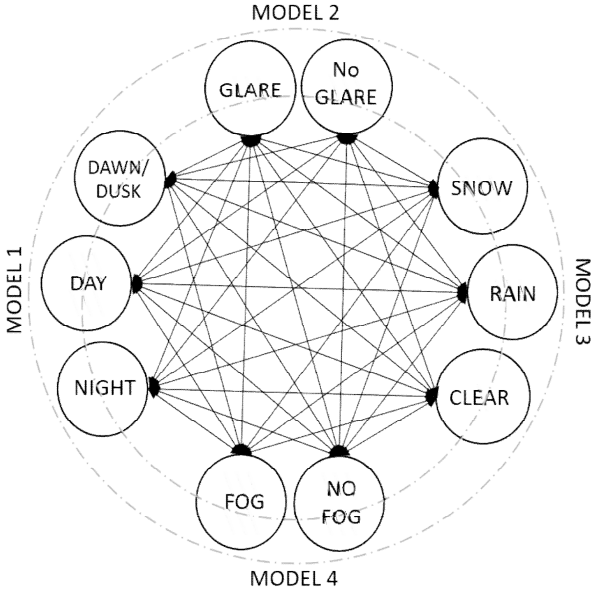

Figure 3: Exclusive vs co-existing classification classes

## 3. METHODOLOGY

### 3.1. WeatherNet framework

To address the current knowledge gap, we introduce a framework of parallel deep CNN models to recognise weather and visual conditions from street-level images of urban scenes, so-called WeatherNet (See Figure 2). This WeatherNet comprises four deep CNN models to detect dawn/dusk, day, night-time, glare, rain, snow, and fog respectively. These models are: 1) NightNet detects the differences between dawn/dusk, day and night-time. It aims to understand the subtleties of street-level images despite the dynamics of weather conditions and urban structure, 2) GlareNet detects images with glare regardless of its source (sun or artificial light) for both dawn/dusk, day and night-time of various weather conditions. Glare is defined as a direct light source that can be seen to cause rings or star effect on the length of the camera without any correction, 3) PreciptationNet detects clear, rainy, or snowy weather for both day and night-time, 4) FogNet detects the occurrence of fog for dawn/dusk, day and night-time and whether this fog happens in the existence of clear, snowy, or rainy weather.

Models 2 and 4 are trained as binary classifiers (0, 1) that detect whether one of the aforementioned events occurs, whereas models 1 and 3 are trained to output one of three classes. The main reasons for training different sets of CNN models then combining them in a framework are the complexity of the classification of urban scenes and the mutual occurrence of one or more of the events at the same time. Figure 3 explains the classes that may occur in one scene by solid arrows, whereas the classes that are mutually exclusive are not linked. For instance, it may be rainy and foggy, during the daytime, while glare is present. Therefore, combining separate models that tackle a certain event in a binary fashion would give a better description of the events in a single image, in addition to the simplicity of the usage and the integration of these models, entirely or partially, for various studies depending on which factors are useful. On the other hand, this binary format makes the precision of the individual models independent from each other, which could allow the modification or improvement of one classifier or more without changing the entire framework.

The training and testing images are resized to (224 X 224 X 3) and fed-forward to the input layer of ResNet50 via transfer learning. Apart from the depth of the architecture of the ResNet, which makes this network a robust one for various classification tasks compared to the previous network is the concept of residual learning. For a further explanation for architecture and the hyperparameters of the model, see (He et al., 2015). The gradients, pre-trained on the ImageNet database (Krizhevsky et al., 2012; Russakovsky et al., 2015), of the different residual blocks of convolution, pooling, batch normalisation layers are set to false, whereas the gradient of the two fully-connected layers of 64 nodes are activated by a ReLU function (Dahl et al., 2013; Glorot et al., 2011), defined as:

$$f(x) = \max(0, x) \qquad (1)$$

where $x$ is the value of the input neuron.

The output layer of the model gives a binary output of single neurons activated based on a sigmoid function, defined as:

$$\delta(x) = \frac{1}{1+e^{-x}} \qquad (2)$$

where $x$ is the value of the input neuron.

The four CNN models are trained based on the back-propagation of error with a batch size of 32, with 'adam' optimiser (Kingma and Ba, 2014) and with an initial learning rate of 0.001 and momentum of 0.9. Each model is trained for 100 training cycles (epochs).

### 3.2 Data

Table 1: Sample size and categories of the data sets

| CNN model | Dataset classes | Sample size |
|---|---|---|
| Model1- NightNet | Dawn/Dusk | 1673 |
| | Day | 2584 |
| | Night | 1848 |
| Model2- GlareNet | Glare | 1159 |
| | No glare | 3549 |
| Model3- PrecipitationNet | Clear | 4017 |
| | Rain | 2343 |
| | Snow | 2347 |
| Model4- FogNet | Fog | 718 |
| | No fog | 3627 |

While Google Street-view images are a good source for various deep learning applications in cities, the images presented there only represent urban areas at a single weather condition, commonly clear weather, neglecting other visual and weather conditions that impact the appearance of cities. On the other hand, there are different datasets for detecting different weather conditions. For instance, the Image2Weather dataset consists of more than 180,000 images of global landmarks of four weather categories, such as sunny, cloudy, rainy, snowy and foggy (Chu et al., 2016) However, the images used for training are still limited and represent cities during the daytime. Accordingly, data augmentation techniques have been applied to enhance the training of each model. The datasets are augmented by rescaling, shearing, horizontal flips, and zooming. These techniques are often common approaches for best practice to enhance the training process and the overall performance of deep learning models (Goodfellow et al., 2017; LeCun et al., 2015).

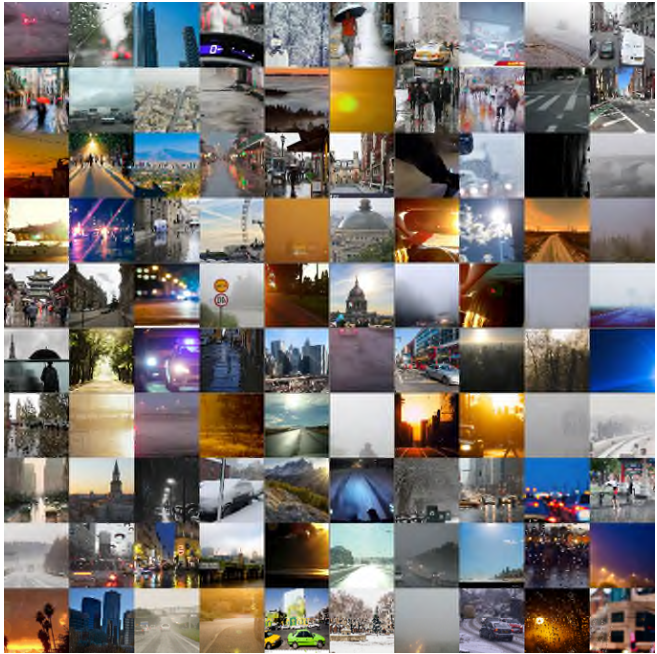

Figure 4: Samples of WeatherNet dataset

Similarly, the Multi-class Weather Image (MWI) dataset consists of 20,000 images of different weather conditions (Z. Zhang et al., 2016). Another example is a binary weather dataset that contains 10,000 images belonging to either sunny or cloudy weather (Lu et al., 2017). Also, a large dataset of images is presented to describe weather conditions from the view of cloud intensity, such as clear, partly cloudy, mostly cloudy, or cloudy, including time and location data (Islam et al., 2013). However, the dataset only represents cities at daytime for clouds intensity neglecting the other factors.

Put all together, creating our own datasets that represent the different environmental conditions of urban scenes is the only way to conduct this research. We have downloaded 23,865 images from the web, specifically Google images for training and testing, using different queries for each class of the weather and visual conditions that includes day and night-time, glare, fog, rain, snow and clear weather. After inspecting these images qualitatively and disregarding images that clearly do not belong to any of the categories, the images are labelled according to each visual class. Table 1 summarises the classes and sample size of data sets used for each model. Figure 4 shows a sample of the multi-class images for training, testing, and validation.

The datasets for each CNN model are subdivided into training and testing sets in 80%-20% train-to-test fashion.

3.3 Evaluation metrics

We evaluate the performance of each CNN model using the following metrics: A cost function of Cross-Entropy to evaluate the model loss during training, testing, and validation. It is defined as:

$$E = -\sum_i^n t_i \log(y_i) \qquad (3)$$

where $t_i$ is the target vector, $y_i$ is the output vector, n represents the number of classes. We also calculated accuracy, precision and recall, false-positive rate, and F1-score for each model, defined as:

$$Accuracy = (TP + TN)/(TP + TN + FP + FN) \qquad (4)$$
$$Precision = TP/(TP + FP) \qquad (5)$$
$$Recall = TP/(TP + FN) \qquad (6)$$
$$False - positive\ rate = FP/(FP + TN) \qquad (7)$$
$$F1 - score = 2 \times \frac{Precision\ X Recall}{Precision + Recall} \qquad (8)$$

Where $TP$ are the predicted true-positive values, $TN$ are the predicted true-negative values, $FP$ are the predicted false-positive values, and $FN$ are the predicted false-negative values.

Last, we compare the performance of our framework with other benchmarks in terms of scope and accuracy. This discussion is partly qualitative due to the absence of benchmark data sets to compare all methods results. However, we also evaluate the performance of WeatherNet on two available datasets, (Gbeminiyi Oluwafemi and Zenghui, 2019; Zhao et al., 2019), and compare the results of our framework with the original outputs.

## 4. RESULTS

Putting all the algorithms of WeatherNet together, the framework can enable the users to extract information of georeferenced weather, and visual conditions to be used for multi-purpose research related to scene awareness, in which weather and visual conditions play a crucial role.

Table 2 summarises the evaluation metrics of each CNN model at the testing phase. After training the four CNN models for 100 epochs, the accuracies for the NightNet, GlareNet, PrecipitationNet, and FogNet on the test data sets are 91.6%, 94.8%, 93.2%, and 95.6% respectively. The models also show high precision and F1-score with low false-positive rates of 6% or lower.

Table 2: Diagnoses of the CNN models for the test sets

| CNN model | Loss (Cross Entropy) | Accuracy (%) | Precision [a] | Recall/ True-positive rate [a] | False-positive rate [a] | F1-score |
|---|---|---|---|---|---|---|
| Model1- NightNet | 0.098 | 91.6 | 0.885 | 0.825 | 0.045 | 0.854 |
| Model2- GlareNet | 0.040 | 94.8 | 0.883 | 0.895 | 0.035 | 0.889 |
| Model3- PrecipitationNet[b] | 0.077 | 93.2 | 0.959 | 0.932 | 0.068 | 0.947 |
| Model4- FogNet | 0.037 | 95.6 | 0.862 | 0.829 | 0.022 | 0.845 |

[a] The metrics are evaluated for the referenced class -indexed zero- for each model.
[b] This model contains three classes, in which the false-positive rate is shared with the classes prior to the referenced class.

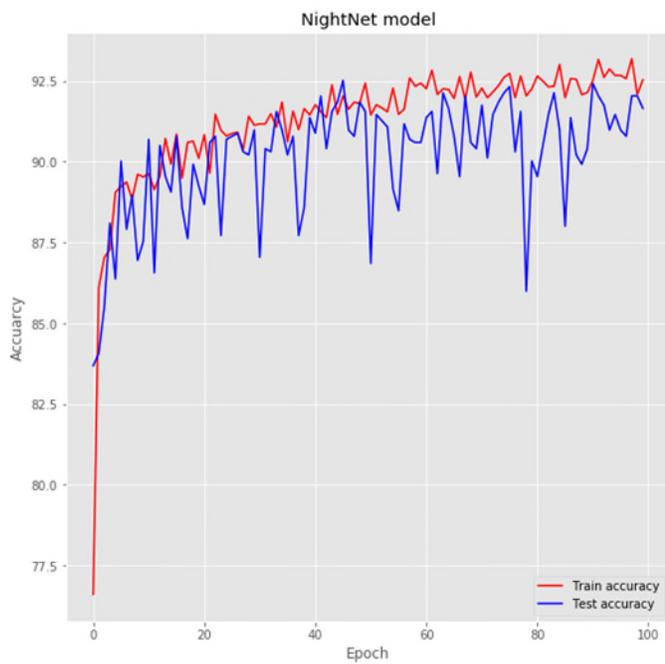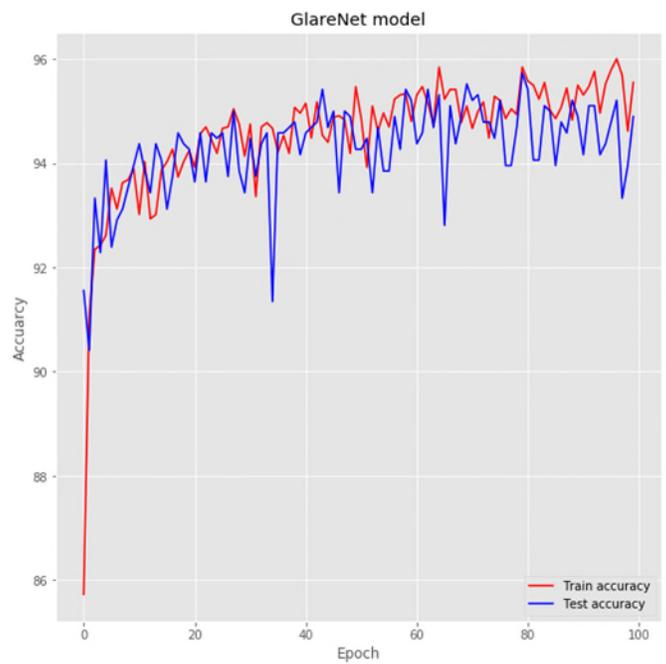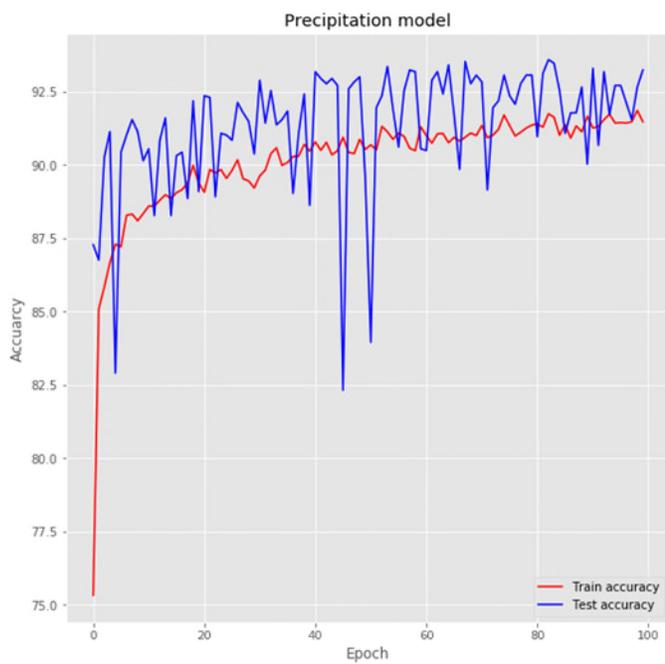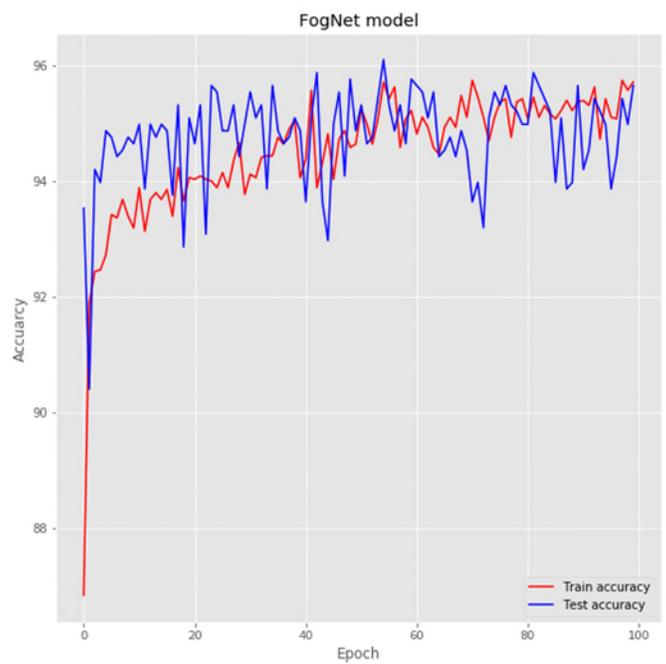

Figure 6: The training and test accuracies per training cycle for each CNN model

In order to investigate further the performance and the fitness of each model during training and testing, figure 6 shows the training and validation accuracies for each epoch, highlighting the overall performance and fitness of each model. It shows the consistency of the accuracies between the training and testing curves, in which no over-fitting is observed. However, due to the high variance in data and subtle differences among classes for the same model, the output for each training cycle show a high level of instability to converge and reach global minimum loss.

Table 3 evaluates our framework against other existing methods that deal with some aspect of weather and visual condition detection. The method used for each model and the yielded accuracy on the dataset used for each paper are also shown. WeatherNet performs favourably in terms of accuracy compared with the other methods, but it should be noted that the datasets used are not the same.

To provide a quantitative comparison, we apply the WeatherNet to two open-sourced datasets used in previous studies (Gbeminiyi Oluwafemi and Zenghui, 2019; Zhao et al., 2019). Table 4 describes the datasets used for evaluation, in terms of size, labels and the original approach used for prediction. The outcomes and evaluation of our model scores on these datasets in comparison to the original are shown, using the same evaluation metrics used in the original research (Accuracy for the first dataset, precision and recall for the second dataset). In case of the first dataset, WeatherNet outperforms the method used for prediction, whereas in the second dataset, our models show higher precision than the original method used for many classification labels, such as clear and rain, and a higher recall for fog and snow detections. For the remaining classes, comparable performance is shown.

Last, as we aim to use the proposed framework pragmatically for recognising and mapping weather and visual conditions in cities, figure 7 shows a few examples of the different model predictions of a wide range of urban scene images taking from different cities globally. It highlights the diversity of the images used for prediction. Regardless of the change of urban structure, camera angles, scene lighting, and components, the proposed models show high accuracy for scene awareness related to visual and weather conditions.

Table 3: Evaluations of the state-of-the-art models based on model types, scope and classification labels

| Methods | | Night-time detection (classes) | Glare detection | Fog detection | Weather detection (classes) | Overall score |
|---|---|---|---|---|---|---|
| (Roser and Moosmann, 2008) | Regions of interest-Histograms | - | - | - | X (clear, light rain, heavy rain) | 0.85 |
| (Islam et al., 2013) | Support Vector Regressor | - | - | - | X (clear, partly cloudy, mostly cloudy, cloudy) | NA |
| (Chu et al., 2016) | Random Forest Classifier | - | - | X | X (Sunny, cloudy, rainy, snowy) | 0.70 |
| (Lu et al., 2017) | CNN model | - | - | - | X (Sunny, cloudy) | 0.91 |
| (Guerra et al., 2018) | Different types of CNN models | - | - | X | X (snowy, rainy) | 0.80 |
| (Gbeminiyi Oluwafemi and Zenghui, 2019) | SAID ENSEMBLE METHOD | - | - | - | X (sunny, cloudy, rainy) | 0.86 |
| (Zhao et al., 2019) | CNN-LSTM | - | - | X | X (sunny, cloudy, rainy, snowy) | 0.91 |
| **WeatherNet** | **Multiple Residual deep models** | **X (Dawn/dusk, day, night)** | **X** | **X** | **X (Clear, rain, snow)** | **0.93** |

Table 4: Evaluations of WeatherNet framework on other open-sourced datasets

| Open-sourced benchmark datasets | | Total images | Labels | Method | Testing scope | Original method score | **WeatherNet score** |
|---|---|---|---|---|---|---|---|
| Multi-class Weather Dataset for Image Classification | (Gbeminiyi Oluwafemi and Zenghui, 2019) | 1,125 | Cloudy, sunshine, rain, sunset | SAID ENSEMBLE METHOD II | Rain detection | Accuracy: 95.20% | **Accuracy: 97.69%** |
| Multi-label weather dataset (test-set) | (Zhao et al., 2019) | 2,000 | (Sunny, cloudy, rainy, snowy, foggy) | CNN-Att-ConvLSTM | Sunny/clear detection Fog detection Rain detection Snow detection | (Precision/Recall): 0.838/0.843 (Precision/Recall): 0.856/0.861 (Precision/Recall): 0.856/0.758 (Precision/Recall): 0.894/0.938 | **(Precision/Recall): 0.924/0.827 (Precision/Recall): 0.833/0.940 (Precision/Recall): 0.958/0.651 (Precision/Recall): 0.789/1** |

Figure 7: The results of the CNN models on street-level images from different cities globally

## 5. DISCUSSION

### 5.1 What makes the WeatherNet state-of-the-art

Cities are complex systems by nature, in which the dynamics of their appearance is highly influenced by multiple factors. Weather and visual conditions are one of these prominent factors that not only impact the appearance of cities but also, complicate the process of understanding them. In this paper, we introduced the WeatherNet framework to tackle the variations and the dynamics of cities' appearances from the perspective of weather and visual aspects. From a single street-level image of an urban scene, the framework is able to capture information related to visual conditions such as dawn-dusk, day or night time, in addition to detecting glare. While on the other hand, the framework can detect weather conditions such as clear, fog, rain, and snow. Figure 8 shows a sample of testing images of various urban settings, visual and weather conditions.

The innovation of the WeatherNet, in comparison to the current state-of-the-art, can be seen in three aspects:
1. The framework is capable of tackling various weather and visual states, including detecting glare, which has never been tackled in any previous deep learning and computer vision research. By using a unified and simple method, the WeatherNet framework is capable of classifying day or night-time, glare, fog, rain, and snow. Most of the previous models recognise only a limited number of weather conditions, neglecting other vital factors.
2. Unlike the current weather recognition models, the proposed framework does not require any pre-defined constraints such as applying filters, defining a camera angle, or defining an action area to the processed image. This simplicity of input makes the proposed framework user-friendly and a base for practical applications for both computer scientists and non-computer scientists to capture information related to weather and appearance of cities from user-defined datasets of street-level images.
3. Although weather and visual conditions depend on time and space, there are no weather stations in each location in cities, and the data forecasted and captured rather represent the agglomeration of locations rather than a precise condition for each location. This undermines the dynamics of the visual appearance of cities. Accordingly, the proposed framework captures weather and visual information. This can enable city planners to map the dynamics of cities according to their weather and visual appearance, which can be a useful tool to understand the dynamics of the appearance of locations and the impacts of these weather and visual dynamics to other aspects of cities (i.e. understanding locations in cities that most likely to cause accidents or risks under certain weather and visual conditions).

### 5.2 Limitations

The proposed framework shows novelty in analysing a wide range of street-level images of cities that belong to various urban structure, visual, and weather conditions globally. The precision of the framework in classification depends on the individual accuracy of each trained CNN model. While the miss-classification error for each classifier is below 8% on the test-sets, in this paper, we only aim to introduce the concept of WeatherNet without further fine-tuning for the CNN hyperparameters or introducing a new architecture that may give better results. As for future work, more experiments with different architectures of CNN models or the way the framework is pipelined may enhance the accuracy.

Compare the performance of the conducted models to previous work remains a limitation due to the absence of weather datasets that comprise similar classes as presented in this paper (i.e. including images of weather at day and night-time and images with and without glare). However, this makes the proposed model indispensable in responding to the current knowledge gap in this research area, and for analysing the variations of urban scene images by deep learning and computer vision that may be helpful for driver-assistance systems or planner and policy-makers in cities.

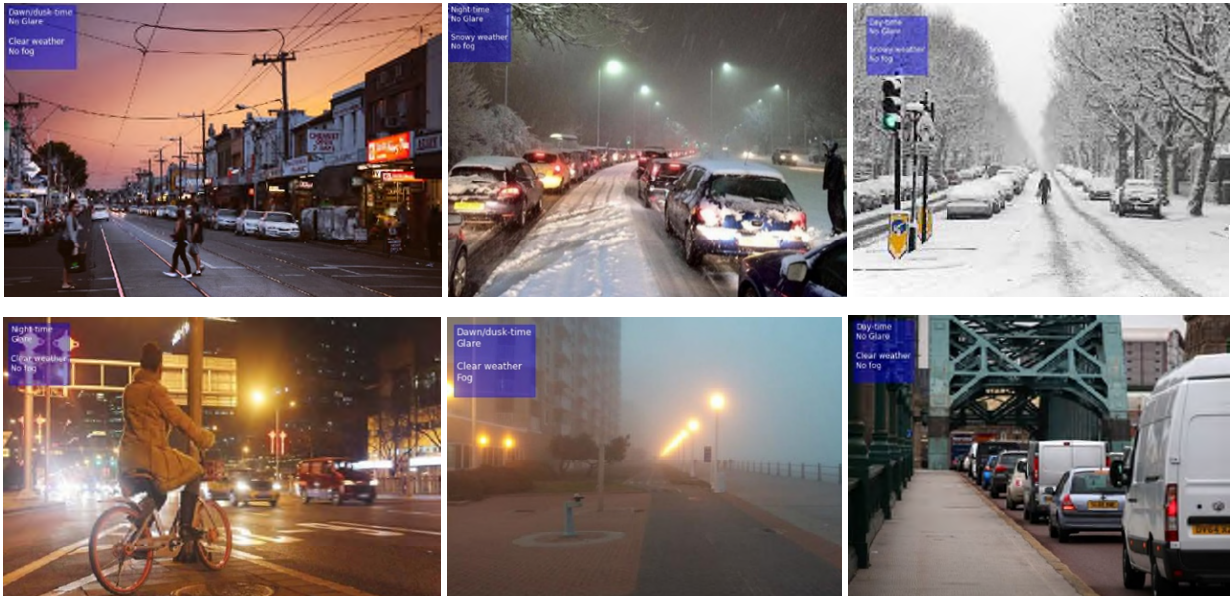

Figure 8: Samples of testing images using the WeatherNet framework

## 6. REMARKS AND FUTURE WORK

In this paper, we present a novel framework, WeatherNet, to detect and map weather and visual conditions from single-images relying on deep learning and computer vision. WeatherNet is capable of detecting 10 classes: Dawn/dusk, day, night, glare, no glare, fog, no fog, clear, rainy, and snowy weather. We aimed to exemplify the application of deep learning and computer vision for scene-awareness and understanding the dynamics of the appearance of urban scenes that could be useful for autonomous applications in cities or elsewhere.

After training four deep CNN models on street-level images from different corners of the globe of various urban structure, weather conditions, and visual appearances, the proposed WeatherNet shows a strong performance in recognising the combination of different categories of a single image. For example, by using WeatheNet framework, urban scenes of street-level images can be classified with multiple classes for a given space and time, such as *'image at daytime, with fog and no rain, in which glare exists'*. The novelty of the proposed framework is in its simplicity for practical applications and for tackling various conditions, in a binary fashion, relying on a unified method without pre-defined constraints for processing images. The proposed framework can be utilised for various proposes; it may be helpful for data automation and autonomous driving in cities, also, it may be utilised towards data automation for mapping and urban planning purposes.

## ACKNOWLEDGMENT

This research outcome is a part of a PhD study for the first author at University College London. This work was supported by UCL Overseas Research Scholarship (ORS) and the Road Safety Trust (RST 38_03_2017). We would like to thank NVIDIA for the GPU grant.